\documentclass{vgtc}                          

\graphicspath{{figures/}{pictures/}{images/}{./}} 

\usepackage{times}                     

\usepackage{tabu}                      
\usepackage{booktabs}                  
\usepackage{lipsum}                    
\usepackage{mwe}                       
\usepackage{amsmath}
\usepackage{mathptmx}                  
\usepackage{array}
\usepackage{enumitem}

\onlineid{0}

\vgtccategory{Research}

\vgtcinsertpkg

\title{AI Scientist Mission Control (AIMC): \\ Visual Analytics for Human Oversight of Autonomous Scientific Discovery}

\author{Rikathi Pal\thanks{e-mail: ripal@cs.stonybrook.edu} %
\and Klaus Mueller\thanks{e-mail:mueller@cs.stonybrook.edu} %
}
\affiliation{\scriptsize Stony Brook University}

\teaser{
  \centering
  \includegraphics[width=\linewidth]{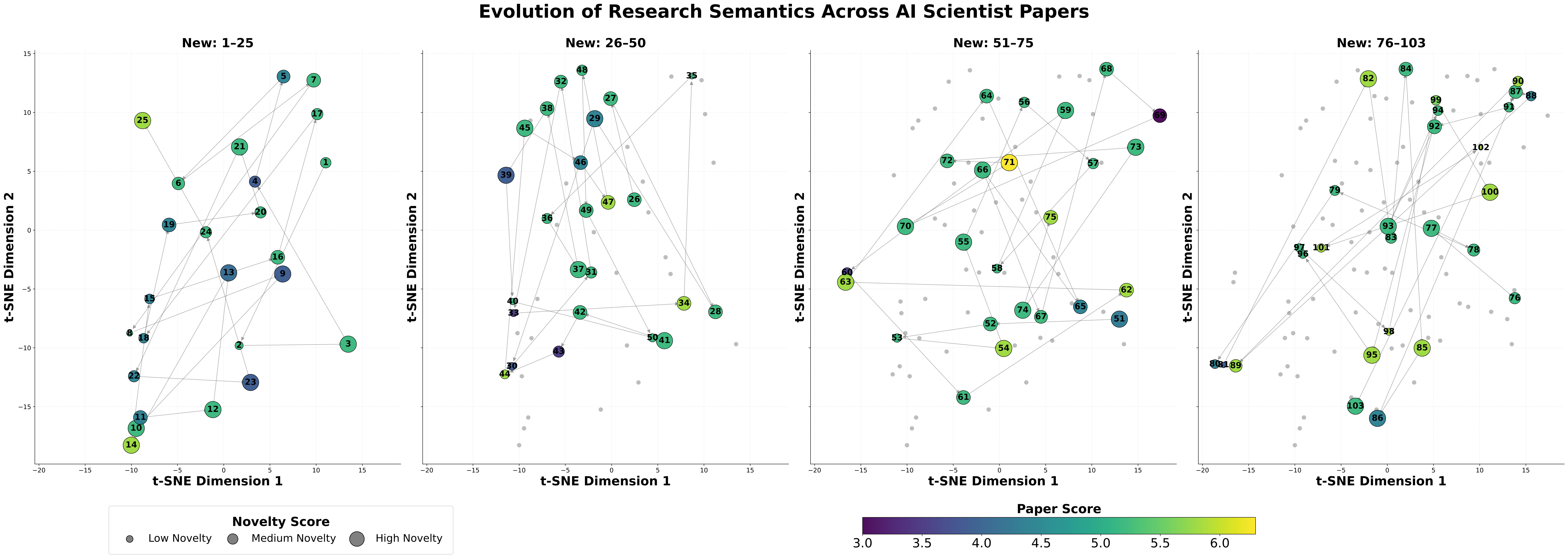}
\caption{Semantic evolution of an autonomous AI Scientist's research landscape. Read from left to right, the four panels show the semantic evolution of AI-generated papers across successive generation intervals (papers 1–25, 26–50, 51–75, and 76–103). Papers are positioned by semantic similarity, colored by automated review score, and sized by within-corpus novelty. Colored nodes represent newly generated papers in each interval, while gray nodes provide the historical context of earlier papers. Later papers increasingly occupy previously sparse semantic regions, suggesting a research frontier that combines the exploitation of established themes with the exploration of new scientific directions.}
  \label{fig:teaser}
}

\abstract{
    Autonomous scientific discovery systems can generate large numbers of research ideas, experiments, and manuscripts with minimal human intervention. As these systems become increasingly capable, scientists require effective mechanisms to monitor output quality, identify recurring failure modes, understand research evolution, and prioritize promising discoveries for review. We present \textbf{AIMC}, a visual analytics framework for human oversight of autonomous scientific discovery. AIMC combines semantic embeddings, automated weakness extraction, temporal analysis, and interactive visualizations to support the exploration of AI-generated research artifacts. We demonstrate the framework through a case study of the papers generated by an autonomous AI Scientist (FARS) \cite{analemma2026fars}, together with their associated review feedback. Our analysis reveals recurring methodological weaknesses, evolving research themes, domain-specific differences in quality, and a small set of highly novel papers that warrant deeper human inspection. These findings illustrate how visual analytics can support transparency, diagnosis, and human AI collaboration in emerging autonomous scientific discovery workflows.

} 

\keywords{Visual Analytics, Autonomous Scientific Discovery, Human Oversight, Human-AI Collaboration, AI Scientists.}

\begin{document}



\firstsection{Introduction}



\maketitle

Recent advances in large language models (LLMs) and agentic AI systems have accelerated the development of autonomous scientific discovery platforms capable of generating research ideas, designing experiments, conducting evaluations, and writing complete research papers with minimal human intervention. 
Systems such as The AI Scientist~\cite{lu2024ai} and the
Fully Automated Research System (FARS)
\cite{analemma2026fars} have demonstrated end-to-end autonomous
research workflows capable of generating hypotheses, conducting
experiments, writing manuscripts, and evaluating research outputs.
As illustrated in Figure~\ref{fig:teaser}, these research outputs collectively form an evolving semantic landscape, reflecting how an autonomous AI scientist revisits existing topics, explores new research directions, and produces papers of varying quality and novelty over time.
While these developments offer exciting opportunities for accelerating research, they also introduce new challenges for human oversight. An AI Scientist can generate hundreds of research artifacts in a short period of time, making manual inspection increasingly impractical. Scientists must understand what research is being produced, identify recurring weaknesses, monitor how research directions evolve, and determine which outputs warrant deeper human review. 

Existing work primarily focuses on improving the capabilities of autonomous scientific discovery systems, including idea generation, experimentation, and manuscript production. However, comparatively little attention has been given to tools that help human researchers oversee the large collections of artifacts these systems generate. In particular, existing approaches do not provide an integrated way to examine research quality, recurring methodological weaknesses, semantic evolution, and review priorities at the corpus level. AIMC addresses this gap by providing a unified visual analytics framework for monitoring, diagnosing, and prioritizing AI-generated scientific research.

We present AI Scientist Mission Control (AIMC), a visual analytics framework for corpus-level oversight of autonomous scientific discovery systems. Rather than developing another AI Scientist, AIMC provides a human-centered analytics layer that enables scientists to monitor, interpret, and evaluate AI-generated research. The framework integrates semantic embeddings, automated weakness extraction, temporal analysis, and quality assessment within coordinated visualizations that reveal recurring failure modes, evolving research themes, domain-specific performance, and promising papers for expert review. The novelty of AIMC lies not in any individual visualization technique, but in integrating semantic, temporal, quality, and methodological signals into a unified oversight framework for autonomous scientific discovery. Unlike traditional literature-analysis tools, AIMC treats generated research artifacts as evidence of an AI Scientist's evolving behavior and connects corpus-level monitoring with diagnosis and human-review prioritization.

To demonstrate the utility of AIMC, we apply it to a corpus of AI-generated papers produced by the Fully Automated Research System (FARS) together with their associated reviewer feedback and evaluation scores. Our case study reveals recurring methodological weaknesses, evolving research agendas, and domain-specific differences in research quality. Although demonstrated using FARS, AIMC is model-agnostic and operates on the outputs of autonomous scientific discovery systems rather than their underlying architectures. Together, these contributions establish visual analytics as a practical approach for monitoring, interpreting, and prioritizing AI-generated scientific research at scale.

The primary contributions of this work are:
\begin{itemize} [leftmargin=*,itemsep=1pt]
    \item  We introduce AIMC, a visual analytics framework that complements existing AI Scientist systems by providing a human-centered oversight layer for monitoring, diagnosing, evaluating, and prioritizing AI-generated scientific research.
    \item We demonstrate how coordinated visual analytics transform large collections of AI-generated papers into actionable insights that support human oversight and decision-making at the corpus level.
    \item Through a case study of the Fully Automated Research System (FARS), we show how visual analytics can reveal recurring failure modes, evolving research agendas, domain-level performance, and high-impact discoveries, informing future improvements to autonomous scientific discovery systems.
\end{itemize}



\section{The FARS Research Corpus}

Our evaluation is based on the Fully Automated Research System (FARS) \cite{analemma2026fars,fars_video}, a publicly deployed autonomous scientific discovery platform that performs end-to-end research generation through a multi-agent workflow encompassing idea generation, experimental design, implementation, evaluation, and manuscript writing~\cite{analemma2026fars,fars_video}. Beginning on February 12, 2026, FARS operated continuously for more than 417 hours, autonomously generating over 166 research papers together with source code, intermediate artifacts, and independent review reports produced by PaperReview.ai \cite{paperreview2026}. The public availability of these research artifacts makes FARS a realistic and reproducible testbed for studying human oversight of autonomous scientific discovery.
For this work, we analyze a corpus of 103 AI-generated papers together with their associated review scores and reviewer feedback. Our study began while FARS was still actively generating new research papers; consequently, the available corpus represented the first 103 papers of an ultimately larger collection exceeding 166 papers. 

Rather than evaluating the underlying AI Scientist itself, our objective is to investigate how visual analytics can help scientists understand the collective behavior of an autonomous scientific discovery system. To analyze research evolution, we treat the paper identifiers assigned by FARS as a chronological sequence, where Paper~1 represents the earliest generated paper and Paper~103 the most recent paper in our study. This ordering serves as the temporal axis for all evolution analyses. The resulting corpus provides a realistic large-scale setting in which AI-generated research artifacts must be monitored, interpreted, and prioritized for expert review, reflecting the oversight challenges posed by future autonomous scientific discovery systems.

\begin{figure*}[h]
    \centering
    \includegraphics[width=\linewidth]{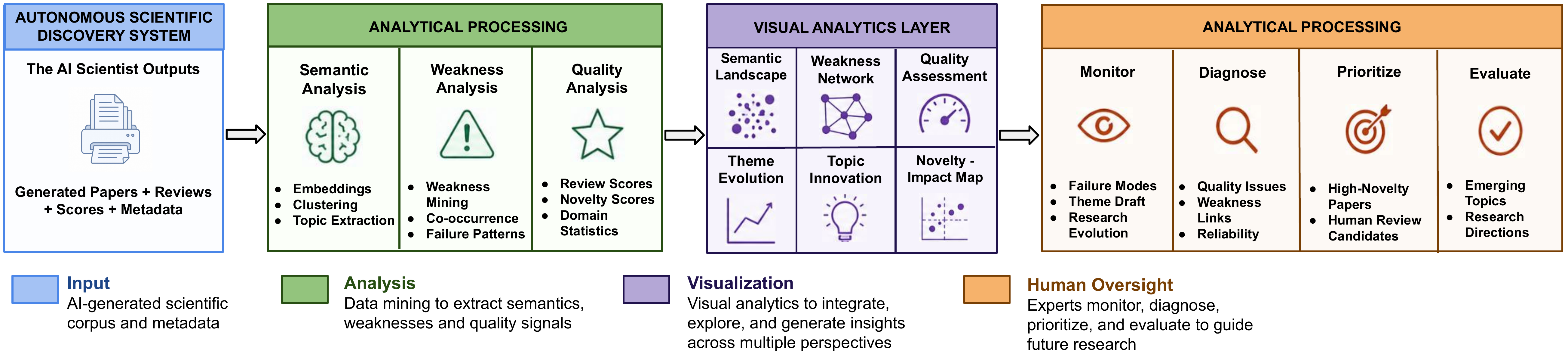}
    \caption{AIMC pipeline for corpus-level oversight of autonomous scientific discovery. AIMC operates on AI-generated papers, reviewer feedback, evaluation scores, and metadata rather than the underlying AI Scientist. Semantic analysis, weakness mining, and quality assessment are integrated into coordinated visualizations that support exploration, diagnosis, evaluation, and prioritization.}
    \label{fig:framework}
\end{figure*}

\section{Related Work}
AIMC lies at the intersection of autonomous scientific discovery and visual analytics. We therefore review related work in three areas: autonomous scientific discovery systems, human oversight of AI-driven science, and visual analytics for human-centered AI. Together, these works highlight the growing need for visual analytics methods that enable scientists to monitor, interpret, and evaluate the outputs of autonomous scientific discovery systems.

\subsection{Autonomous Scientific Discovery}

Recent advances in large language models and agentic AI have enabled autonomous scientific discovery systems capable of generating research ideas, conducting experiments, writing manuscripts, and evaluating scientific outputs. Systems including \textit{The AI Scientist}~\cite{lu2024ai}, \textit{The AI Scientist-v2}~\cite{yamada2025ai}, \textit{Agent Laboratory}~\cite{schmidgall2025agent}, \textit{AI-Researcher}~\cite{tang2026ai}, \textit{MLAgentBench}~\cite{huang2023mlagentbench}, \textit{AutoSciLab}~\cite{desai2025autoscilab}, and \textit{Analemma's Fully Automated Research System (FARS)}~\cite{analemma2026fars}, have substantially advanced this vision through autonomous research agents, benchmarking platforms, and self-driving laboratories~\cite{wei2025ai}. As these systems become more capable, effective oversight of their outputs becomes equally important.

\subsection{Human Oversight of Autonomous Science}

Despite increasing autonomy, scientific reasoning and validation remain fundamentally human-centered. Recent surveys on agentic AI for scientific discovery ~\cite{gridach2025agentic} argue that AI scientists should augment rather than replace human researchers, emphasizing the continued importance of human expertise throughout the scientific process. As autonomous scientific discovery systems become increasingly capable, maintaining transparency, accountability, and effective human oversight is recognized as a key challenge for their responsible deployment~\cite{wei2025ai}.  Holzinger et al. \cite{holzinger2025human} argue that meaningful human oversight is essential for ensuring trustworthy AI systems capable of supporting critical scientific decision-making. However, most existing research focuses on improving autonomous research agents rather than helping scientists inspect, interpret, and steer the large collections of research artifacts the agents generate, motivating visual analytics approaches for corpus-level oversight.
Human oversight remains necessary because autonomous systems can generate large volumes of research containing methodological errors, unsupported claims, or misleading evaluations. Automated reviewers may reproduce similar errors and cannot fully assess scientific significance, contextual validity, or ethical consequences. Human experts are therefore needed to verify important claims, interpret uncertain evidence, and determine which outputs warrant further investigation. AIMC supports this process by directing limited human attention toward potentially important or problematic outputs. Human oversight remains necessary because autonomous systems can generate large volumes of research containing methodological errors, unsupported claims, or misleading evaluations. Automated reviewers may reproduce similar errors and cannot fully assess scientific significance, contextual validity, or ethical consequences. Human experts are therefore needed to verify important claims, interpret uncertain evidence, and determine which outputs warrant further investigation. AIMC supports this process by directing limited human attention toward potentially important or problematic outputs.

\subsection{Visual Analytics for Human-Centered AI}

Visual analytics combines computational analysis with interactive visualization to support human reasoning and decision-making \cite{keim2008visual}. More recently, it has become an important paradigm for improving the transparency and interpretability of AI systems through human-centered exploration of model behavior and explanations \cite{alicioglu2022survey,chatzimparmpas2025visual,rong2023towards}. Human-computer collaboration further emphasizes the integration of human expertise into intelligent analytical workflows \cite{monadjemi2023human}.
Existing literature-visualization tools primarily support the exploration of human-authored publications, whereas AI-monitoring tools generally focus on model performance, execution traces, or individual outputs. In contrast, AIMC analyzes collections of AI-generated research artifacts as evidence of an autonomous system's evolving behavior. It integrates semantic, temporal, quality, and methodological signals to support corpus-level monitoring, diagnosis, and prioritization for expert review. Together, these distinctions motivate the application of visual analytics to human oversight in autonomous scientific discovery and inform the AIMC design goals presented in the following section.

\section{AIMC Design Goals}

Based on the requirements identified in the previous sections, AIMC is designed around five complementary design goals.

\begin{itemize}[leftmargin=*,itemsep=1pt]
\item \textbf{DG1: Integrate Multi-Modal Research Evidence.}
Combine paper content, reviewer feedback, evaluation scores, novelty, and metadata into a unified analytical representation.

\item \textbf{DG2: Support Coordinated Exploration.}
Provide linked visualizations that reveal semantic structure, methodological weaknesses, research quality, and temporal evolution.

\item \textbf{DG3: Reveal Corpus-Level Behavior.}
Expose recurring failure modes, thematic trends, and quality patterns that emerge only across large collections of AI-generated research.

\item \textbf{DG4: Enable Human Oversight.}
Support monitoring, diagnosis, and prioritization to help scientists efficiently identify outputs requiring expert attention.

\item \textbf{DG5: Inform Continuous Improvement.}
Facilitate longitudinal analysis of research evolution to inform future refinement of autonomous scientific discovery systems.
\end{itemize}

These goals are realized through coordinated visualizations, each supporting a complementary oversight task, summarized in Table~\ref{tab:tasks}.

\begin{table}[t]
\centering
\caption{AIMC analytical views and supported oversight tasks.}
\label{tab:tasks}
\small
\resizebox{\linewidth}{!}{%
\begin{tabular}{
>{\centering\arraybackslash}p{3cm}
>{\centering\arraybackslash}p{2.9cm}
>{\centering\arraybackslash}p{4cm}
}
\toprule
\textbf{Analytical View} &
\textbf{Primary Data} &
\textbf{Oversight Objective} \\
\midrule
Quality Distribution &
Review scores &
Assess overall research quality \\

Weakness Network &
Reviewer feedback &
Diagnose recurring failure modes \\

Weakness Comparison &
Reviewer feedback &
Prioritize critical weaknesses \\

Weakness Evolution &
Temporal reviews &
Monitor reliability over time \\

Theme Evolution &
Semantic embeddings &
Track research agenda shifts \\

Domain Comparison &
Domain + scores &
Compare performance across domains \\

Novelty-Impact Map &
Novelty + review score &
Prioritize high-value discoveries \\
\bottomrule

\end{tabular}%
}
\end{table}

\section{AIMC Framework}

Figure~\ref{fig:framework} provides an overview of the AIMC pipeline. The framework operates on the outputs of an autonomous scientific discovery system, including AI-generated papers, reviewer feedback, evaluation scores, and associated metadata. These heterogeneous inputs form the scientific corpus that is processed through three complementary analytical modules. AIMC is independent of the architecture or generation process of any particular AI Scientist. It can be applied to other autonomous scientific discovery platforms provided that they produce research artifacts, temporal or provenance metadata, and assessment signals such as review scores, textual feedback, or human annotations. The analytical modules are adaptable: different embedding models, domain-specific weakness taxonomies, and alternative quality measures can be substituted without changing the overall oversight workflow. AIMC can therefore support cross-platform analysis while accommodating differences in artifact formats, scientific domains, and evaluation procedures.

The first module performs \textit{semantic analysis}, where paper embeddings, clustering, and topic extraction are used to capture semantic relationships, reveal thematic structure, and characterize the evolution of research topics. The second module performs \textit{weakness analysis}, mining reviewer feedback to identify recurring methodological weaknesses, co-occurring failure modes, and systematic patterns across the generated corpus. The third module performs \textit{quality analysis}, integrating review scores, novelty estimates, and domain statistics to assess research quality and impact.
The extracted analytical signals are then integrated within a coordinated visual analytics layer comprising multiple complementary views, including semantic landscapes, theme evolution, weakness networks, topic innovation, quality assessment, and novelty-impact analysis. Rather than presenting these analyses independently, the coordinated views enable scientists to explore relationships across multiple perspectives and obtain a corpus-level understanding of AI-generated research.
These visualizations support four complementary human oversight tasks. Scientists can \textit{monitor} research quality, recurring failure modes, and research evolution; \textit{diagnose} methodological weaknesses and their relationships; \textit{prioritize} high-novelty papers and candidates for expert review; and \textit{evaluate} emerging research topics and broader research directions. Together, these capabilities transform large collections of AI-generated research into actionable insights that enhance transparency, support human decision-making, and establish a foundation for future human-in-the-loop autonomous scientific discovery. AIMC does not independently validate scientific claims or replace expert review; instead, it helps researchers prioritize artifacts and patterns that require closer human examination.

\textbf{Usage scenario.} A researcher can begin with the quality-distribution view to assess the corpus and identify papers requiring attention. The researcher can then use the weakness views to diagnose recurring methodological problems and the temporal and domain views to determine where these problems persist. Finally, the novelty--impact view helps prioritize promising or uncertain papers for expert inspection. In this way, AIMC connects multiple analytical views into a unified workflow from corpus-level monitoring to targeted human review.

\section{Case Study}

\begin{figure}
    \centering
    \includegraphics[width=0.9\linewidth]{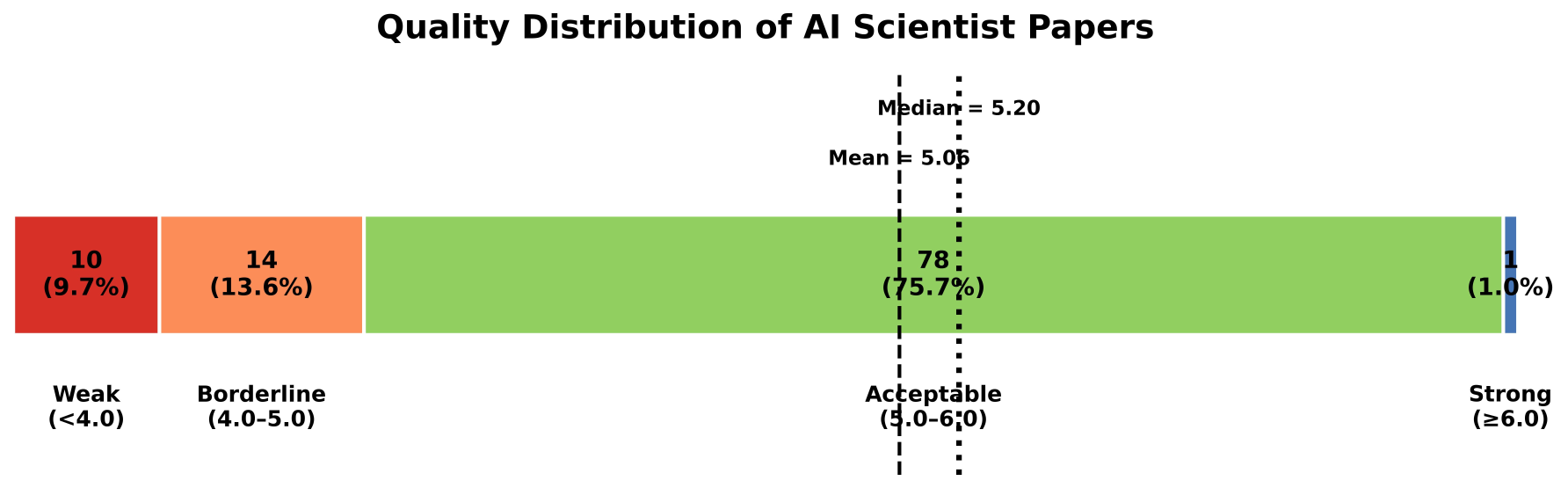}  
    \caption{Review-score distribution of FARS-generated papers based on the PaperReview.ai evaluations available through the FARS website. Most papers (75.7\%) fall within the Acceptable range (5.0–6.0), with very few receiving Strong ratings ($\geq$6.0). The concentration of scores around the mean and median suggests that the system consistently produces research of acceptable quality. 
    }
    \label{fig:score_distribution}
    \vspace{- 10 pt}
\end{figure}

To demonstrate the utility of AIMC, we present a case study based on research artifacts generated by the Fully Automated Research System (FARS)~\cite{analemma2026fars}. Although our analysis is based on FARS, AIMC is model-agnostic and applicable to other AI Scientist systems that generate research artifacts together with associated evaluation metadata.
Rather than evaluating individual papers in isolation, our goal is to understand the broader behavior of an autonomous scientific discovery system. 
Specifically, we investigate four questions that are central to human oversight: 
\begin{enumerate}[leftmargin=*,nosep]
\item What recurring failure modes characterize AI-generated research?
\item How does the research agenda of the AI Scientist evolve over time?
\item Which research domains produce stronger or weaker outputs?
\item Which generated papers warrant deeper human review?
\end{enumerate}


The following analyses explain how AIMC helps scientists answer these questions through visual analytics.

\subsection{Baseline Quality Assessment}

Human oversight begins with an overall assessment of research quality before investigating individual failure modes. The quality distribution of the papers (Figure~\ref{fig:score_distribution}), as assessed by PaperReview.ai \cite{paperreview2026}, provides a high-level overview of the review score distribution across the AI-generated corpus, enabling scientists to quickly assess the quality profile of the generated research.

The predominance of moderately scored papers suggests that the primary oversight challenge is not only to identify poor outputs, but also to efficiently direct human attention within the large collection of generally acceptable research artifacts. This baseline assessment motivates our subsequent analyses, which examine recurring weaknesses, research evolution, domain-specific performance, and novelty to identify papers and patterns warranting closer inspection.
This finding can guide researchers to allocate more review effort to borderline papers rather than inspecting the entire corpus uniformly.

\begin{figure}
    \centering
    \includegraphics[width=0.9\linewidth]{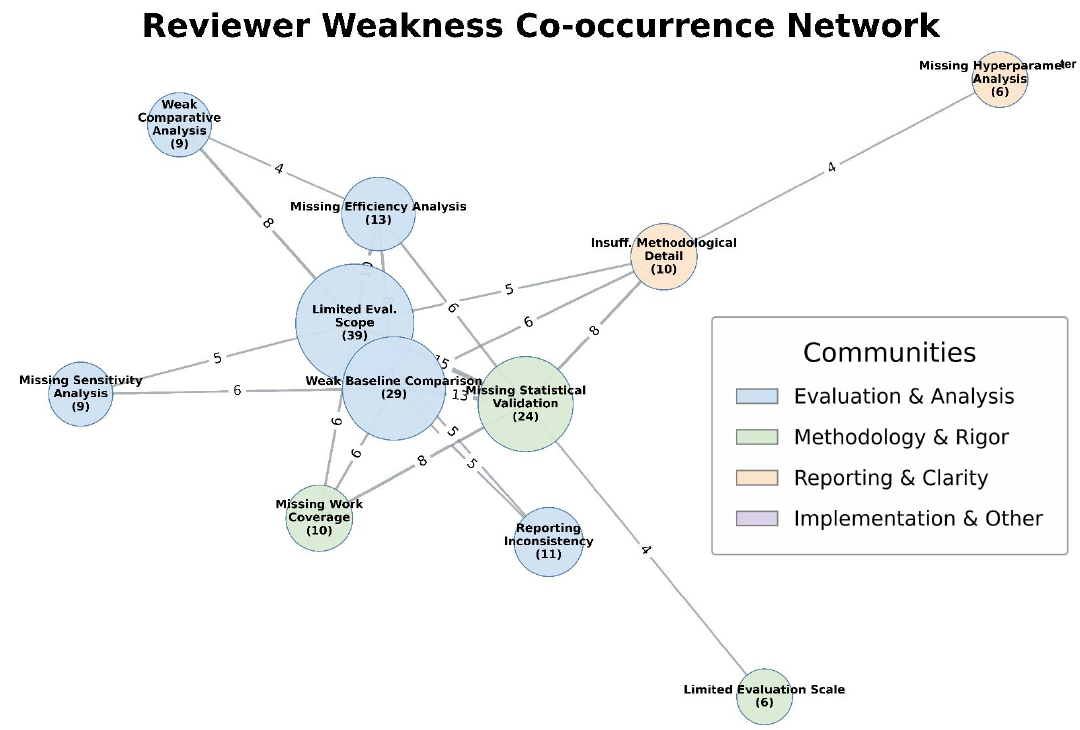}  
    \caption{
Reviewer weakness co-occurrence network for AI-generated papers. Node size represents the frequency of each weakness across the corpus, while edge thickness and labels denote the number of papers in which two weaknesses co-occur. Distinct communities reveal groups of methodological weaknesses that frequently occur together.
}
    \label{fig:weakness_network}
\end{figure}

\subsection{What Characterizes Recurring Failure Modes?}

A central objective of human oversight is to understand why autonomous scientific discovery systems repeatedly produce lower-quality research. To support this objective, AIMC provides three complementary analyses that identify recurring methodological weaknesses, quantify their relationship to research quality, and track how these failure modes evolve over time. Reviewer-identified weaknesses were automatically extracted from the free-text review reports using the OpenAI GPT API, then manually verified and standardized into a consistent set of methodological weakness categories for subsequent analysis.

\textbf{Weakness Co-occurrence.} We used a mass spring algorithm to visualize the reviewer weakness co-occurrence network (Figure~\ref{fig:weakness_network}). The visualization reveals that methodological deficiencies rarely occur in isolation. Instead, weaknesses organize into interconnected communities associated with evaluation, methodological rigor, and reporting quality issues, indicating that lower-quality papers often exhibit multiple related shortcomings simultaneously.

\begin{figure}
    \centering
    \includegraphics[width=\linewidth]{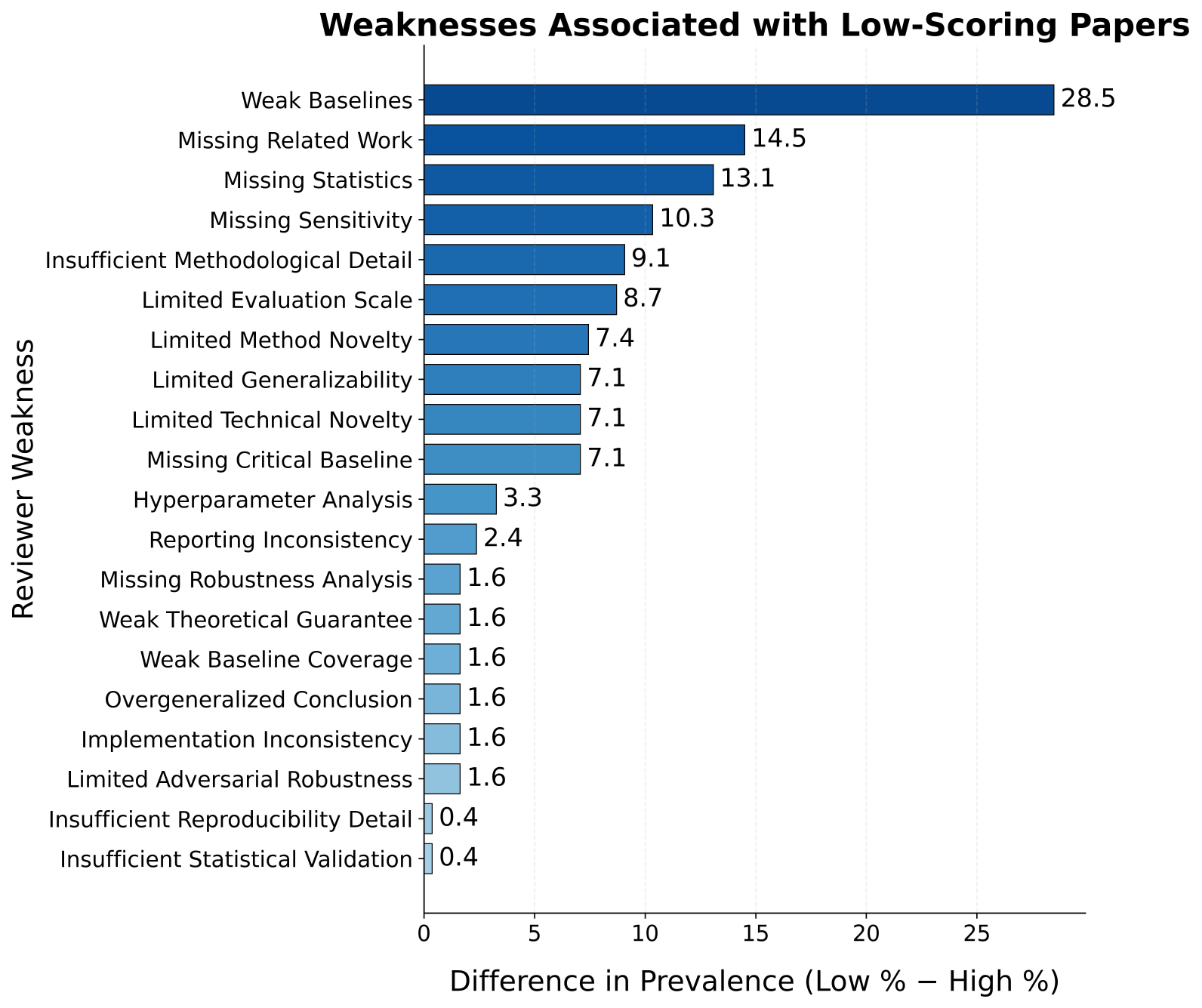}
        \vspace{-15 pt}
    \caption{Differences in reviewer-identified weaknesses between low-scoring (review score $<$ 5.0) and high-scoring (review score $\geq$ 5.0) AI-generated papers. Bars represent prevalence differences (Low\% $-$ High\%), with larger values indicating weaknesses more strongly associated with lower-rated papers.
    }
    \label{fig:low_high_weaknesses}
    \vspace{-0 pt}
\end{figure}

\textbf{Weaknesses Associated with Lower-Quality Papers.} Building on this structural view, Figure~\ref{fig:low_high_weaknesses} compares the prevalence of reviewer criticisms between low-scoring (review score $<$ 5.0) and high-scoring (review score $\geq$ 5.0) papers. Rather than displaying the prevalence of each group separately, each horizontal bar represents the difference in prevalence (Low\% $-$ High\%) for a given weakness. Consequently, longer and darker blue bars indicate weaknesses that occur substantially more frequently in low-scoring papers, whereas shorter and lighter blue bars indicate smaller differences between the two groups. The analysis reveals that weak baseline comparisons, missing related work, insufficient statistical validation, and limited sensitivity analyses are disproportionately associated with lower-quality research. Together with the weakness co-occurrence network, these findings suggest that methodological rigor—particularly experimental design and evaluation—rather than novelty is the primary factor distinguishing stronger AI-generated papers from weaker ones.

\begin{figure} [t!]
    \centering
    \includegraphics[width=\linewidth]{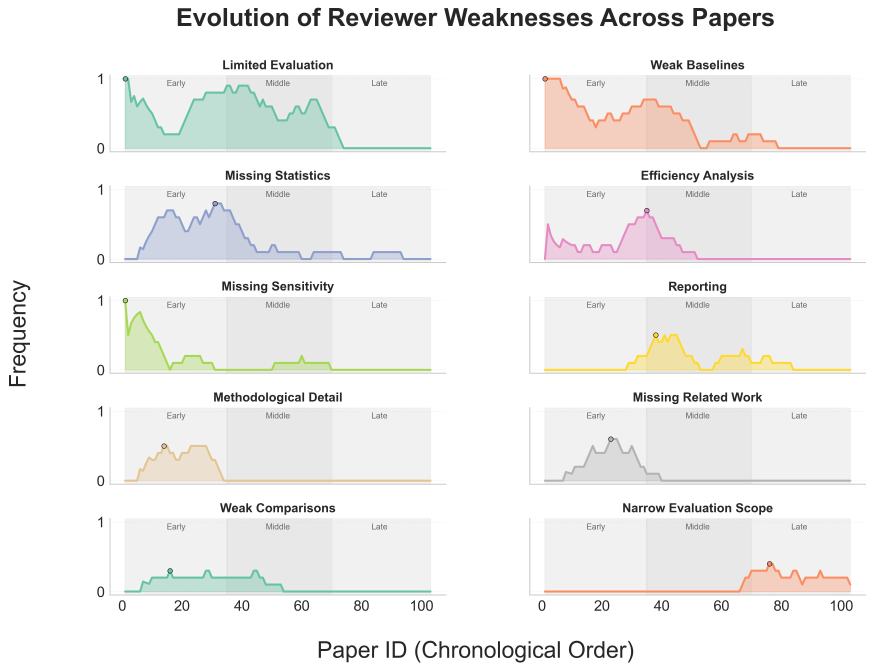}
            \vspace{-15 pt}
    \caption{    
    Evolution of reviewer-identified weaknesses across the AI Scientist's research timeline. Rolling frequencies show that some methodological weaknesses become less frequent over time, whereas others, particularly those related to evaluation, become more prevalent. Here, \textbf{Limited Evaluation} denotes insufficient validation of the proposed method, whereas \textbf{Narrow Evaluation Scope} denotes evaluation restricted to a single task, dataset, or domain.
    }
        \vspace{-10 pt}
    \label{fig:weakness_evolution}
\end{figure}

\textbf{Evolution of Reviewer Weaknesses.} While the first two analyses identify recurring weaknesses and their impact on research quality, effective oversight also requires understanding whether these deficiencies change as the AI Scientist continues generating research. Figure~\ref{fig:weakness_evolution}, therefore tracks the temporal evolution of common reviewer-identified weaknesses across the chronological sequence of generated papers. The results show that several weaknesses, including methodological detail and weak comparisons, gradually become less frequent, whereas others, such as narrow evaluation scope, emerge during later stages of generation. This demonstrates that the AI Scientist's failure modes evolve alongside its research agenda rather than remaining static. 

Together, this set of visualizations transforms reviewer feedback into actionable diagnostic knowledge. Rather than identifying isolated problems, AIMC enables scientists to uncover systematic failure patterns, determine which weaknesses most strongly contribute to lower research quality, and monitor how these limitations change over time. These insights support targeted human intervention and guide the iterative refinement of autonomous scientific discovery systems and workflows.
Researchers can use these findings to prioritize papers with recurring critical weaknesses and identify methodological areas requiring system refinement.

\subsection{How Does the Research Agenda Evolve Over Time?}

\begin{figure}[t]
\centering
\includegraphics[width=\linewidth]{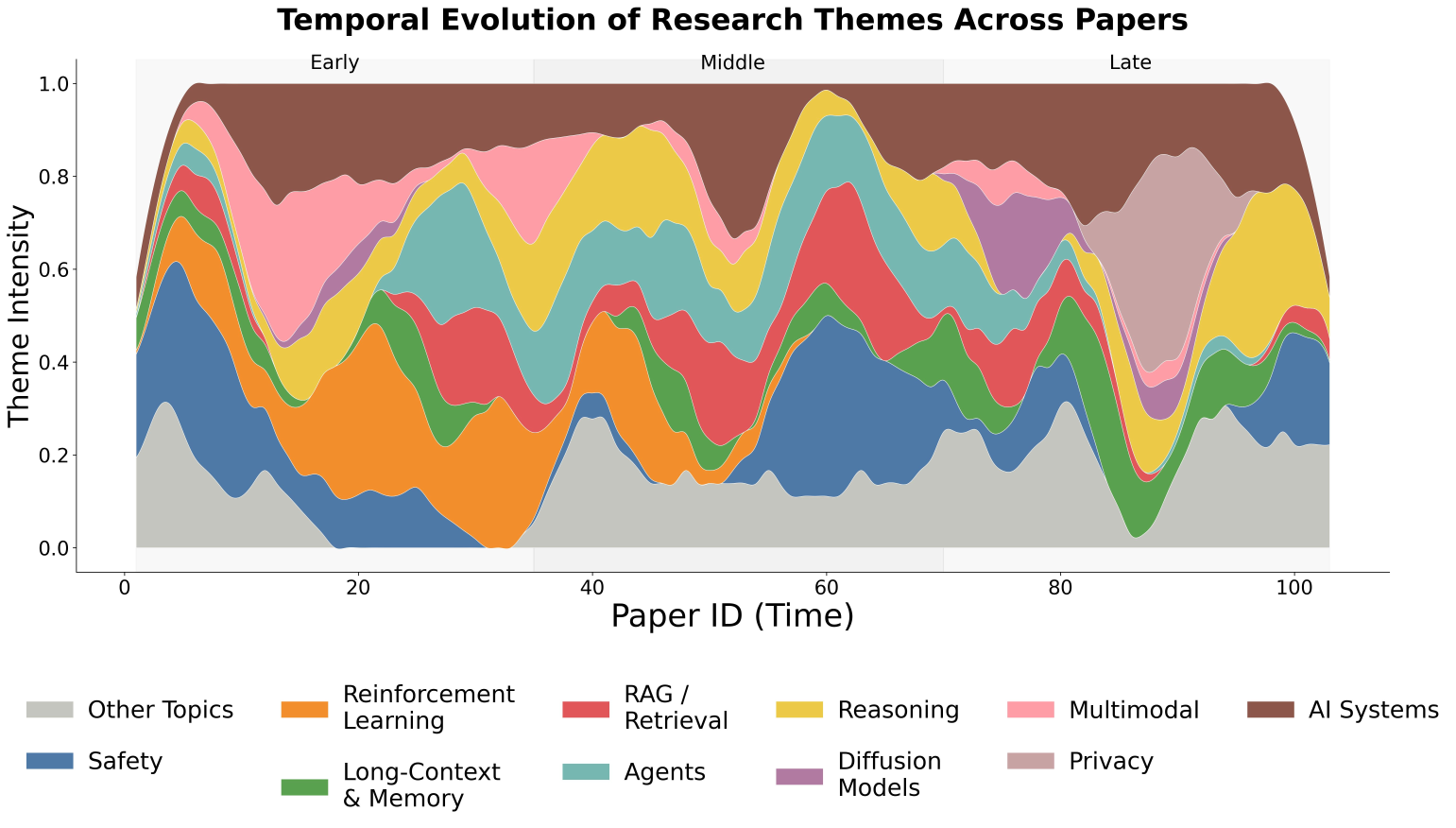}
\caption{Temporal evolution of research themes across AI-generated papers. The stacked area chart shows how the relative prominence of research themes changes over time, revealing shifts in the AI Scientist's research focus.
}
\label{fig:theme_evolution}
    \vspace{-10 pt}
\end{figure}

\begin{figure}[t]
\centering
\includegraphics[width=\linewidth]{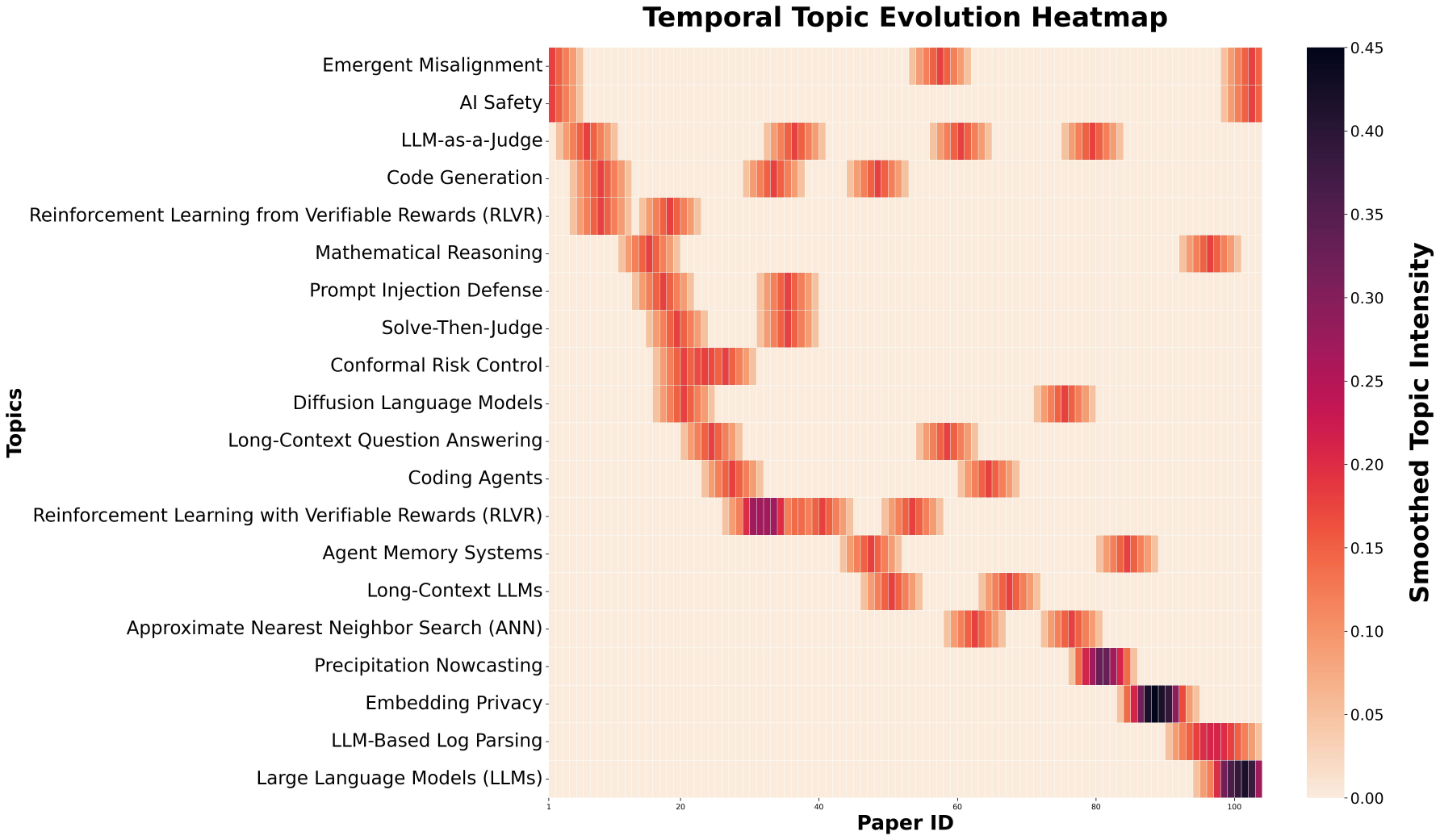}

\caption{Topic evolution heatmap of the AI Scientist's research corpus. Temporal changes in topic prevalence reveal how individual research areas emerge, decline, and occasionally reappear, complementing the broader theme-level analysis of Figure \ref{fig:theme_evolution}. Topic intensities are shown after moving-average smoothing (window size = 10).
}
\label{fig:topic_evolution}
\end{figure}

Beyond evaluating research quality, effective human oversight requires understanding how an autonomous scientific discovery system explores and adapts its research agenda over time. AIMC addresses this question through complementary analyses of research themes and topics (Figures~\ref{fig:theme_evolution} and~\ref{fig:topic_evolution}). Themes were obtained by grouping semantically related paper keywords into broader research categories, while individual keywords were treated as fine-grained topics. Paper IDs served as a proxy for chronological order, and because FARS generated multiple papers concurrently, theme and topic frequencies were smoothed using a centered moving-average filter with a 10-paper window to highlight long-term trends.

\textbf{Temporal Evolution.} The temporal theme evolution view (Figure~\ref{fig:theme_evolution}) provides a high-level perspective on how the AI Scientist allocates its effort across major research directions throughout the generation process. Rather than progressing through a fixed sequence of themes, the system dynamically reallocates its attention among multiple research areas, revisiting established themes while introducing new ones. Interestingly, a persistent fraction of the research effort remains devoted to the \emph{Other Topics} category throughout the corpus, indicating that the AI Scientist continually explores a diverse set of less common research directions rather than concentrating exclusively on its dominant themes.

\textbf{Topic Evolution.} Complementing this corpus-level perspective, the topic evolution heatmap (Figure~\ref{fig:topic_evolution}) provides a finer-grained view of individual research topics. Many topics appear as relatively short-lived bursts of activity before declining or occasionally reappearing, suggesting that the AI Scientist continuously redirects its attention toward new research problems rather than repeatedly generating variations of the same ideas.

\textbf{Multi-Level Research Evolution.} Together, these coordinated visualizations complement the semantic evolution illustrated in Figure~\ref{fig:teaser}. Whereas the teaser depicts the research landscape expanding in semantic space, the temporal analyses reveal the dynamics underlying this process. Broad research themes remain active over extended periods, while individual topics emerge and fade more rapidly. A persistent long tail of less common topics indicates continual exploration beyond the dominant research areas, extending the semantic frontier into previously sparse regions. By combining theme-level trends with topic-level dynamics, AIMC enables scientists to understand how autonomous scientific discovery systems balance sustained investment in established themes with the exploration of emerging scientific directions.
These trends can help researchers decide whether to encourage exploration of emerging topics or strengthen underrepresented research directions.


\subsection{Which Domains Produce Strong or Weak Outputs?}
\begin{figure}
    \centering
    \includegraphics[width=\linewidth]
{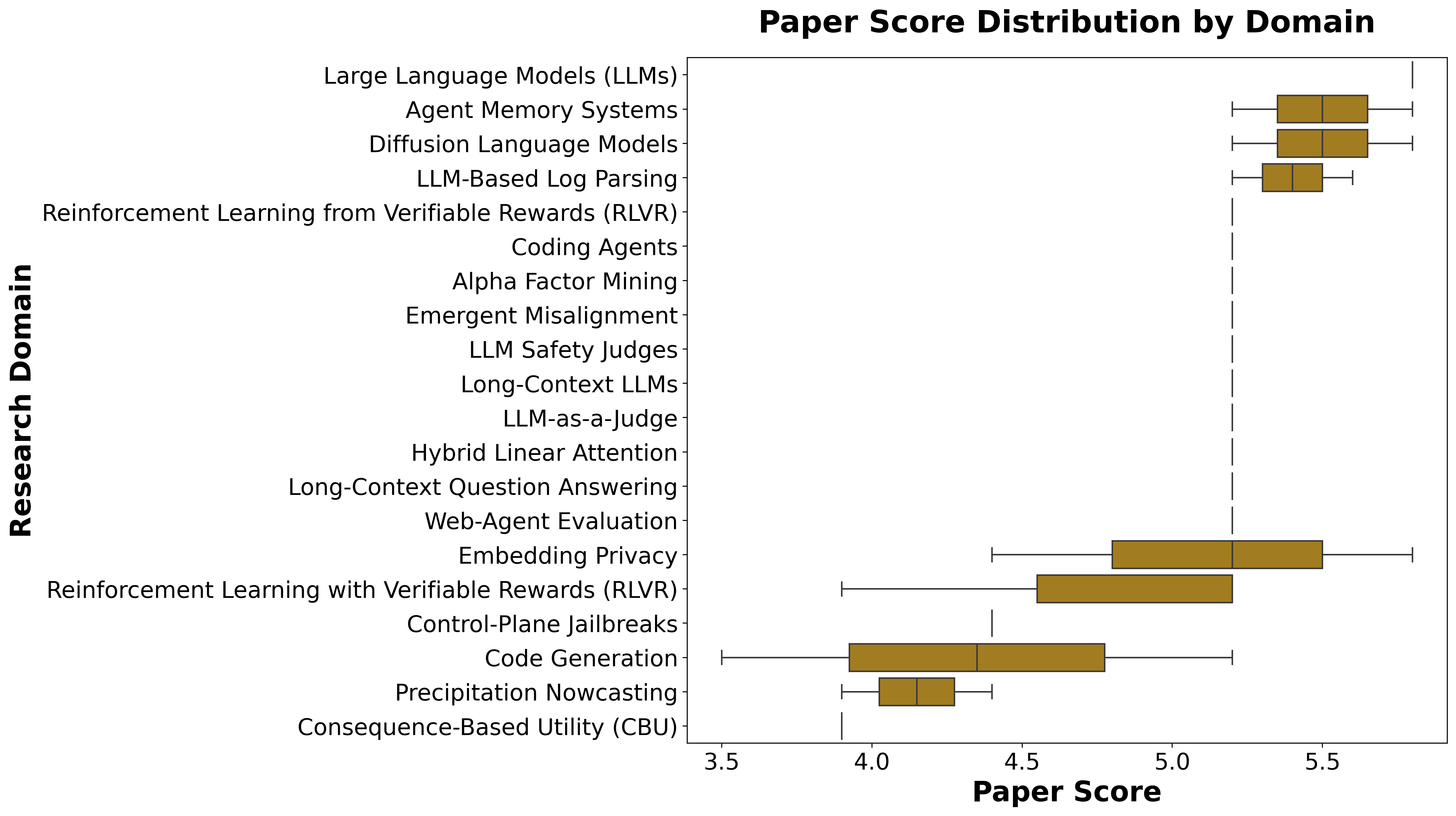}
    \vspace{-10 pt}
    \caption{Quality comparison across research domains. Differences in median scores and score variability suggest that research quality varies across domains, with some domains consistently producing higher-rated papers than others.
    }
    \label{fig:domain_scores}
    \vspace{-10 pt}
\end{figure}

Autonomous scientific discovery systems may perform differently across various research domains, making domain-level evaluation a crucial component of human oversight. AIMC addresses this question by comparing the quality of AI-generated research across different scientific domains (Figure~\ref{fig:domain_scores}).

The domain-level comparison reveals that research quality varies substantially across the AI Scientist's research portfolio. While several domains consistently produce higher-quality outputs with relatively stable review scores, others exhibit lower median performance together with greater score variability, indicating that these research areas are both more challenging and less consistent for the AI Scientist. 

These domain-specific differences illustrate the value of corpus-level oversight. Rather than relying on aggregate performance alone, this analysis exposes strengths and weaknesses that are specific to individual research domains. Scientists can now identify research areas that warrant closer inspection, target methodological refinement where most needed, and monitor whether improvements occur consistently across domains.
The domain-level differences can guide researchers in selecting domains that require additional validation, improved evaluation criteria, or greater expert involvement.




\subsection{Which New Papers Warrant Human Attention?}
\begin{figure}
    \centering
    \includegraphics[width=\linewidth]{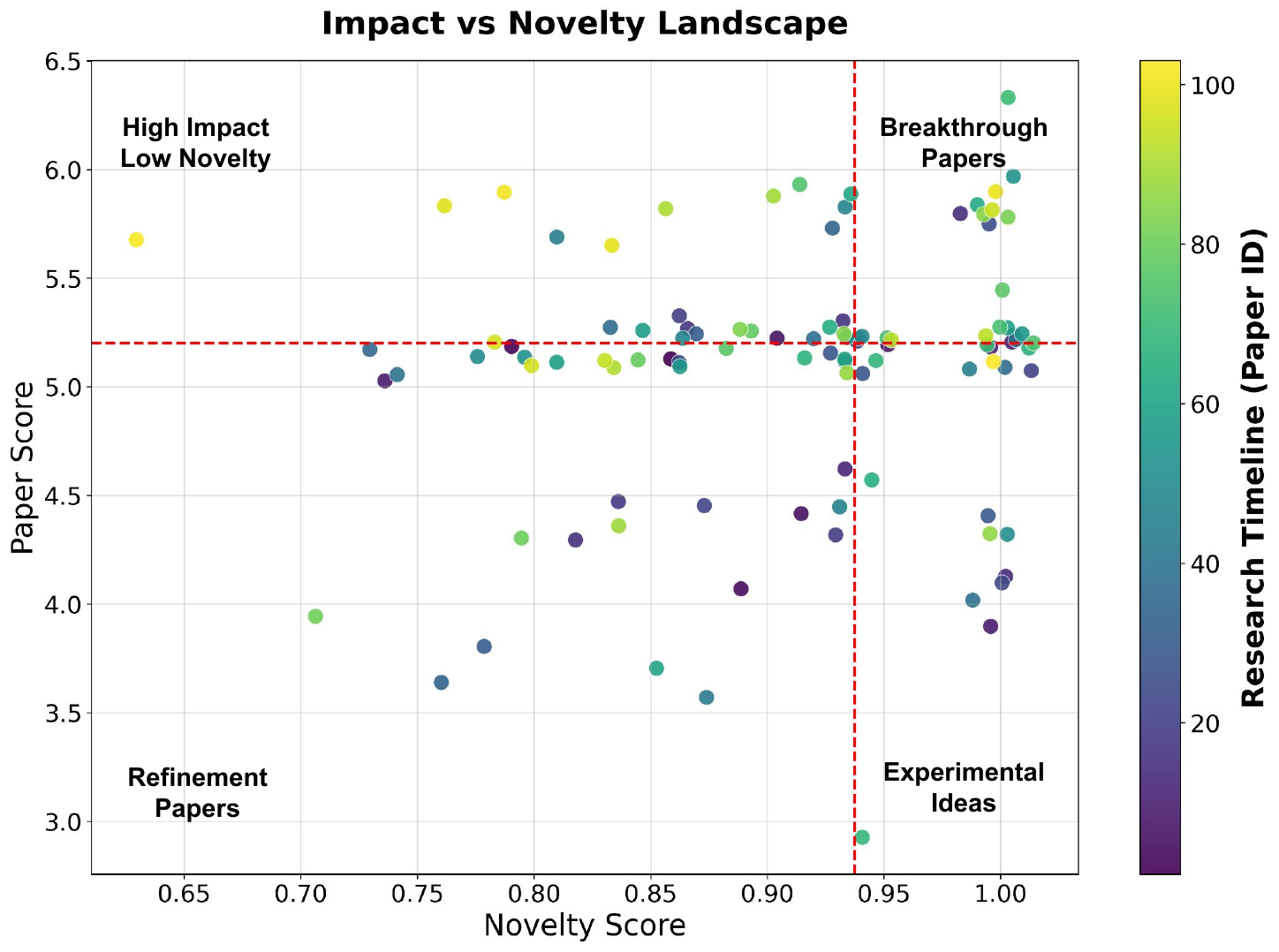}
    \caption{Novelty-impact landscape of AI-generated research papers. Each point represents a paper positioned by its novelty and review scores. Dashed lines indicate the median values, partitioning the space into four prioritization categories. The upper-right quadrant identifies high-novelty, high-impact papers.   
    }
    \label{fig:novelty_impact}
    \vspace{-10 pt}
\end{figure}

As autonomous scientific discovery systems produce research at scales that exceed human review capacity, identifying the most promising papers for expert inspection becomes an important human oversight task. AIMC addresses this challenge by jointly considering the novelty and impact of each generated paper through the novelty-impact landscape shown in Figure~\ref{fig:novelty_impact}.

To estimate \emph{novelty}, we aggregate keywords extracted from the corpus and compute the average inverse keyword frequency ($1/\mathrm{frequency}$) for each paper, assigning higher scores to papers containing relatively rare research concepts. \emph{Impact} is represented by the paper's review score obtained from the AI Scientist evaluation pipeline. The median novelty and impact values partition the visualization into four categories that support paper prioritization.

The novelty-impact landscape reveals that originality and research quality provide complementary perspectives on AI-generated research. While papers in the upper-right quadrant combine high novelty and high impact, making them promising candidates for expert review, the remaining quadrants capture distinct types of contributions. The upper-left quadrant contains well executed but comparatively incremental research, whereas the lower-right quadrant highlights highly novel ideas whose scientific potential is limited by methodological shortcomings. Rather than ranking papers by a single metric, the visualization enables scientists to distinguish between reliable refinements, exploratory ideas, and potentially high-impact discoveries.

The temporal coloring provides an additional perspective on the evolution of the AI Scientist's research. Later papers generally receive higher review scores than earlier ones, suggesting that the overall quality of the generated research improves over the corpus. However, highly novel papers appear throughout the later stages of the timeline rather than being concentrated among the most recent outputs, indicating that improvements in research quality do not necessarily coincide with increasing novelty. The upper-left quadrant is dominated by relatively recent papers that achieve strong review scores despite only moderate novelty, reflecting well-executed incremental contributions, whereas the upper-right quadrant contains a smaller collection of highly novel, high-impact papers distributed across the later stages of the corpus. Interestingly, the presence of a small number of earlier papers in this quadrant suggests that breakthrough contributions can emerge throughout the generation process rather than only after prolonged exploration. Together, these observations demonstrate that novelty and research quality capture complementary aspects of AI-generated research and reinforce the value of jointly considering novelty, impact, and temporal context when allocating limited human review effort.
Researchers can use this view to prioritize high-impact and corpus-distinctive papers for expert review while flagging unusual but weakly evaluated ideas for further verification.

\section{Discussion \& Conclusion}
Autonomous scientific discovery is fundamentally changing how scientific knowledge is generated, shifting researchers from conducting experiments toward supervising increasingly capable AI Scientists. As these systems begin producing research at scales beyond manual inspection, effective human oversight becomes essential for ensuring transparency, reliability, and informed scientific decision-making.

We presented AIMC, a visual analytics framework for human oversight of autonomous scientific discovery. Rather than developing another AI Scientist, AIMC complements existing autonomous research platforms by providing the visual analytics infrastructure needed to understand, audit, and prioritize AI-generated scientific outputs. Through a case study of AI-generated research, we demonstrated how coordinated visualizations enable scientists to assess research quality, diagnose recurring methodological weaknesses, analyze research evolution, compare domain-specific performance, and prioritize promising discoveries. Collectively, these analyses show that evaluating AI Scientists requires more than inspecting individual papers or aggregate benchmark scores. Corpus-level visual analytics reveals behavioral patterns that emerge only across large collections of AI-generated research artifacts.
Uncertainty in AI-generated reviews. AIMC uses automated review scores and feedback, which may contain errors, biases, hallucinations, or domain-specific inconsistencies. Because both the papers and their evaluations are AI-generated, errors from these two stages may compound and produce misleading patterns. These signals should therefore support, rather than replace, expert judgment. Future work will incorporate multiple reviewers, uncertainty measures, and human validation.

One particularly interesting observation concerns the evolution of the AI Scientist's research. While we initially expected the semantic embedding visualization to exhibit progressively stronger clustering as additional papers were generated, the semantic landscape instead expanded into previously sparse regions. Together with the temporal analyses of research themes and topics, this suggests that the AI Scientist continually explores new semantic directions while maintaining established research areas. Similarly, the novelty-impact analysis revealed that later papers generally achieve higher review scores, whereas highly novel contributions emerge throughout the later stages of the corpus rather than only among the most recent papers. Taken together, these findings suggest that research quality and novelty evolve differently, highlighting the importance of considering multiple complementary perspectives when evaluating autonomous scientific discovery systems.

Although AIMC employs visualization techniques commonly used for exploring scientific literature, its objective is fundamentally different. Traditional literature exploration systems analyze human-generated publications, whereas AIMC treats AI-generated research artifacts as observable evidence of an autonomous scientific discovery system's behavior. Rather than studying science itself, the framework studies the behavior of the AI Scientist through the scientific artifacts it produces.

Beyond supporting oversight, AI-generated research corpora may also provide a valuable resource for studying scientific practice itself. Interestingly, many of the dominant reviewer criticisms identified in our analysis, including inadequate baselines, insufficient statistical analysis, and limited methodological detail, closely resemble those commonly raised for human-authored research. This suggests that autonomous scientific discovery systems are subject to many of the same methodological expectations as human researchers. More broadly, AI-generated corpora may provide a new opportunity for meta-scientific analyses that identify recurring methodological shortcomings and help reinforce best practices for conducting and reporting scientific research.

The limitations of our current study are, first, the evaluation is based on a single dataset from one autonomous scientific discovery platform (FARS), which may limit the generalizability of the findings. Second, the keywords and weakness categories were extracted and consolidated automatically and may contain errors or oversimplifications. The automated review scores and feedback may also be unreliable, biased, or inconsistent across domains. In addition, the structure of the t-SNE visualization may vary with the selected embedding model and parameter settings and should not be interpreted as definitive evidence of semantic relationships. Finally, AIMC has not yet been formally evaluated by domain experts, and its usability and analytical findings require human validation.

Despite these limitations, AIMC is model-agnostic and applicable to other AI Scientist systems. Future work will evaluate the framework across multiple platforms, incorporate domain experts, and introduce multiple reviewers and uncertainty measures. We also envision AIMC evolving into a closed-loop human-AI collaboration framework in which visual insights guide the iterative refinement of autonomous scientific discovery systems. Future versions could additionally integrate execution traces, experiment logs, planning artifacts, and intermediate reasoning to combine process-level and outcome-level evidence. As autonomous scientific discovery systems mature, visual analytics can become an essential component of transparent, trustworthy, and collaborative AI-driven scientific discovery.
\bibliographystyle{abbrv-doi}

\bibliography{template}

@misc{analemma2026fars,
  author       = {{Analemma Intelligence}},
  title        = {FARS: Fully Automated Research System},
  year         = {2026},
  howpublished = {\url{https://analemma.ai/fars/}},
  note         = {Accessed: 2026-06-25}
}

@article{lu2024ai,
  title={The AI scientist: Towards fully automated open-ended scientific discovery},
  author={Lu, Chris and Lu, Cong and Lange, Robert Tjarko and Foerster, Jakob and Clune, Jeff and Ha, David},
  journal={arXiv preprint arXiv:2408.06292},
  year={2024}
}

@article{huang2023mlagentbench,
  title={MLAgentBench: Evaluating Language Agents on Machine Learning Experimentation},
  author={Huang, Qian and Vora, Jian and Liang, Percy and Leskovec, Jure},
  journal={arXiv preprint arXiv:2310.03302},
  year={2023},
  eprint={2310.03302},
  archivePrefix={arXiv},
  primaryClass={cs.LG}
}

@misc{fars_video,
  author       = {{Analemma Intelligence}},
  title        = {FARS: Fully Automated Research System (Video)},
  year         = {2026},
  howpublished = {\url{https://www.youtube.com/watch?v=lT8LGvS7Os8}},
  note         = {YouTube video. Accessed: 2026-07-01}
}

@article{schmidgall2025agent,
  title={Agent laboratory: Using llm agents as research assistants},
  author={Schmidgall, Samuel and Su, Yusheng and Wang, Ze and Sun, Ximeng and Wu, Jialian and Yu, Xiaodong and Liu, Jiang and Moor, Michael and Liu, Zicheng and Barsoum, Emad},
  journal={Findings of the Association for Computational Linguistics: EMNLP 2025},
  pages={5977--6043},
  year={2025},
  publisher={Association for Computational Linguistics}
}

@article{yamada2025ai,
  title={The ai scientist-v2: Workshop-level automated scientific discovery via agentic tree search},
  author={Yamada, Yutaro and Lange, Robert Tjarko and Lu, Cong and Hu, Shengran and Lu, Chris and Foerster, Jakob and Clune, Jeff and Ha, David},
  journal={arXiv preprint arXiv:2504.08066},
  year={2025}
}

@article{tang2026ai,
  title={Ai-researcher: Autonomous scientific innovation},
  author={Tang, Jiabin and Xia, Lianghao and Li, Zhonghang and Huang, Chao},
  journal={Advances in Neural Information Processing Systems},
  volume={38},
  pages={9481--9520},
  year={2026}
}

@article{wei2025ai,
  title={From ai for science to agentic science: A survey on autonomous scientific discovery},
  author={Wei, Jiaqi and Yang, Yuejin and Zhang, Xiang and Chen, Yuhan and Zhuang, Xiang and Gao, Zhangyang and Zhou, Dongzhan and Wang, Guangshuai and Gao, Zhiqiang and Cao, Juntai and others},
  journal={arXiv preprint arXiv:2508.14111},
  year={2025}
}

@inproceedings{desai2025autoscilab,
  title={AutoSciLab: A self-driving laboratory for interpretable scientific discovery},
  author={Desai, Saaketh and Addamane, Sadhvikas and Tsao, Jeffrey Y and Brener, Igal and Swiler, Laura P and Dingreville, Remi and Iyer, Prasad P},
  booktitle={Proceedings of the AAAI Conference on Artificial Intelligence},
  volume={39},
  number={1},
  pages={146--154},
  year={2025}
}

@article{gridach2025agentic,
  title={Agentic ai for scientific discovery: A survey of progress, challenges, and future directions},
  author={Gridach, Mourad and Nanavati, Jay and Abidine, Khaldoun Zine El and Mendes, Lenon and Mack, Christina},
  journal={arXiv preprint arXiv:2503.08979},
  year={2025}
}

@article{holzinger2025human,
  title={Is human oversight to AI systems still possible?},
  author={Holzinger, Andreas and Zatloukal, Kurt and M{\"u}ller, Heimo},
  journal={New Biotechnology},
  volume={85},
  pages={59--62},
  year={2025},
  publisher={Elsevier}
}

@article{rong2023towards,
  title={Towards human-centered explainable ai: A survey of user studies for model explanations},
  author={Rong, Yao and Leemann, Tobias and Nguyen, Thai-Trang and Fiedler, Lisa and Qian, Peizhu and Unhelkar, Vaibhav and Seidel, Tina and Kasneci, Gjergji and Kasneci, Enkelejda},
  journal={IEEE Transactions on Pattern Analysis and Machine Intelligence},
  volume={46},
  number={4},
  pages={2104--2122},
  year={2023},
  publisher={IEEE}
}

@article{alicioglu2022survey,
  title={A survey of visual analytics for explainable artificial intelligence methods},
  author={Alicioglu, Gulsum and Sun, Bo},
  journal={Computers \& Graphics},
  volume={102},
  pages={502--520},
  year={2022},
  publisher={Elsevier}
}

@article{chatzimparmpas2025visual,
  title={Visual analytics for explainable and trustworthy artificial intelligence},
  author={Chatzimparmpas, Angelos},
  journal={IEEE Computer Graphics and Applications},
  volume={45},
  number={2},
  pages={100--111},
  year={2025},
  publisher={IEEE}
}

@inproceedings{monadjemi2023human,
  title={Human--Computer Collaboration for Visual Analytics: an Agent-based Framework},
  author={Monadjemi, Shayan and Guo, Mengtian and Gotz, David and Garnett, Roman and Ottley, Alvitta},
  booktitle={Computer Graphics Forum},
  volume={42},
  number={3},
  pages={199--210},
  year={2023},
  organization={Wiley Online Library}
}

@incollection{keim2008visual,
  title={Visual analytics: Definition, process, and challenges},
  author={Keim, Daniel and Andrienko, Gennady and Fekete, Jean-Daniel and G{\"o}rg, Carsten and Kohlhammer, J{\"o}rn and Melan{\c{c}}on, Guy},
  booktitle={Information Visualization: Human-Centered Issues and Perspectives},
  pages={154--175},
  year={2008},
  publisher={Springer}
}

@misc{paperreview2026,
  author       = {{PaperReview.ai}},
  title        = {PaperReview.ai},
  year         = {2026},
  howpublished = {\url{https://paperreview.ai/}},
  note         = {AI-assisted scientific paper review platform. Accessed: July 3, 2026}
}
\end{document}